%% file: iclr2025_arxiv.tex
\documentclass{article} 
\usepackage{iclr2025_conference,times}

\input{math_commands.tex}

\usepackage[utf8]{inputenc} 
\usepackage[T1]{fontenc}    
\usepackage{hyperref}       
\usepackage{url}            
\usepackage{booktabs}       
\usepackage{amsfonts}       
\usepackage{nicefrac}       
\usepackage{microtype}      
\usepackage{xcolor}         
\usepackage{multirow}
\usepackage{graphicx}

\usepackage{amsmath}
\usepackage{amssymb}
\usepackage{cleveref}
\usepackage{wrapfig}
\usepackage{bm}
\usepackage{multirow}
\usepackage{algorithm}
\usepackage[noend]{algpseudocode}
\usepackage{colortbl}

\definecolor{Gray}{gray}{0.8999}

\title{LongHalQA: Long-Context Hallucination Evaluation for MultiModal Large Language Models}

\author{Han Qiu$^1$, Jiaxing Huang$^1$, Peng Gao$^2$, Qin Qi$^2$, Xiaoqin Zhang$^3$, Ling Shao$^4$, Shijian Lu$^{1}$\thanks{Corresponding author.}\\
$^1$S-Lab, Nanyang Technological University, $^2$Shanghai Artificial Intelligence Laboratory,\\
$^3$College of Computer Science and Technology, Zhejiang University of Technology, \\
$^4$UCAS-Terminus AI Lab, UCAS\\
{\tt\small han023@e.ntu.edu.sg, \{Jiaxing.Huang, Shijian.Lu\}@ntu.edu.sg}
}

\iclrfinalcopy 
\begin{document}

\maketitle

\begin{abstract}
Hallucination, a phenomenon where multimodal large language models~(MLLMs) tend to generate textual responses that are plausible but unaligned with the image, has become one major hurdle in various MLLM-related applications. Several benchmarks have been created to gauge the hallucination levels of MLLMs, by either raising discriminative questions about the existence of objects or introducing LLM evaluators to score the generated text from MLLMs. However, the discriminative data largely involve simple questions that are not aligned with real-world text, while the generative data involve LLM evaluators that are computationally intensive and unstable due to their inherent randomness. We propose LongHalQA, an LLM-free hallucination benchmark that comprises 6K long and complex hallucination text. LongHalQA is featured by GPT4V-generated hallucinatory data that are well aligned with real-world scenarios, including object/image descriptions and multi-round conversations with 14/130 words and 189 words, respectively, on average. It introduces two new tasks, hallucination discrimination and hallucination completion, unifying both discriminative and generative evaluations in a single multiple-choice-question form and leading to more reliable and efficient evaluations without the need for LLM evaluators. Further, we propose an advanced pipeline that greatly facilitates the construction of future hallucination benchmarks with long and complex questions and descriptions. Extensive experiments over multiple recent MLLMs reveal various new challenges when they are handling hallucinations with long and complex textual data. Dataset and evaluation code are available at \url{https://github.com/hanqiu-hq/LongHalQA}.
\end{abstract}

\section{Introduction}

Multi-modal Large Language Models~(MLLMs)~\citep{dai2024instructblip,hu2024minicpm,fuyu8b2023,Qwen-VL,llava15,liu2024llavanext,zhu2023minigpt} have achieved great progress in understanding multi-modal contents, by generating detailed descriptions of images, conducting sophisticated, consecutive conversations with humans, etc. Despite the remarkable advancements, MLLMs often experience severe hallucination problems~\citep{yin2023woodpecker,leng2023mitigating,huang2023opera,zhu2024ibd,yue2024less,bai2024hallucination} by generating textual responses that are not aligned with the corresponding image contents. While hallucinations significantly compromise MLLMs' reliability and applicability in various vision-language tasks and applications, effective and efficient measurement of the hallucination level of MLLMs has become a prerequisite for diagnosis and mitigation of hallucination in MLLMs.

\begin{figure}[t]
\footnotesize
\centering
\includegraphics[width=0.97\linewidth]{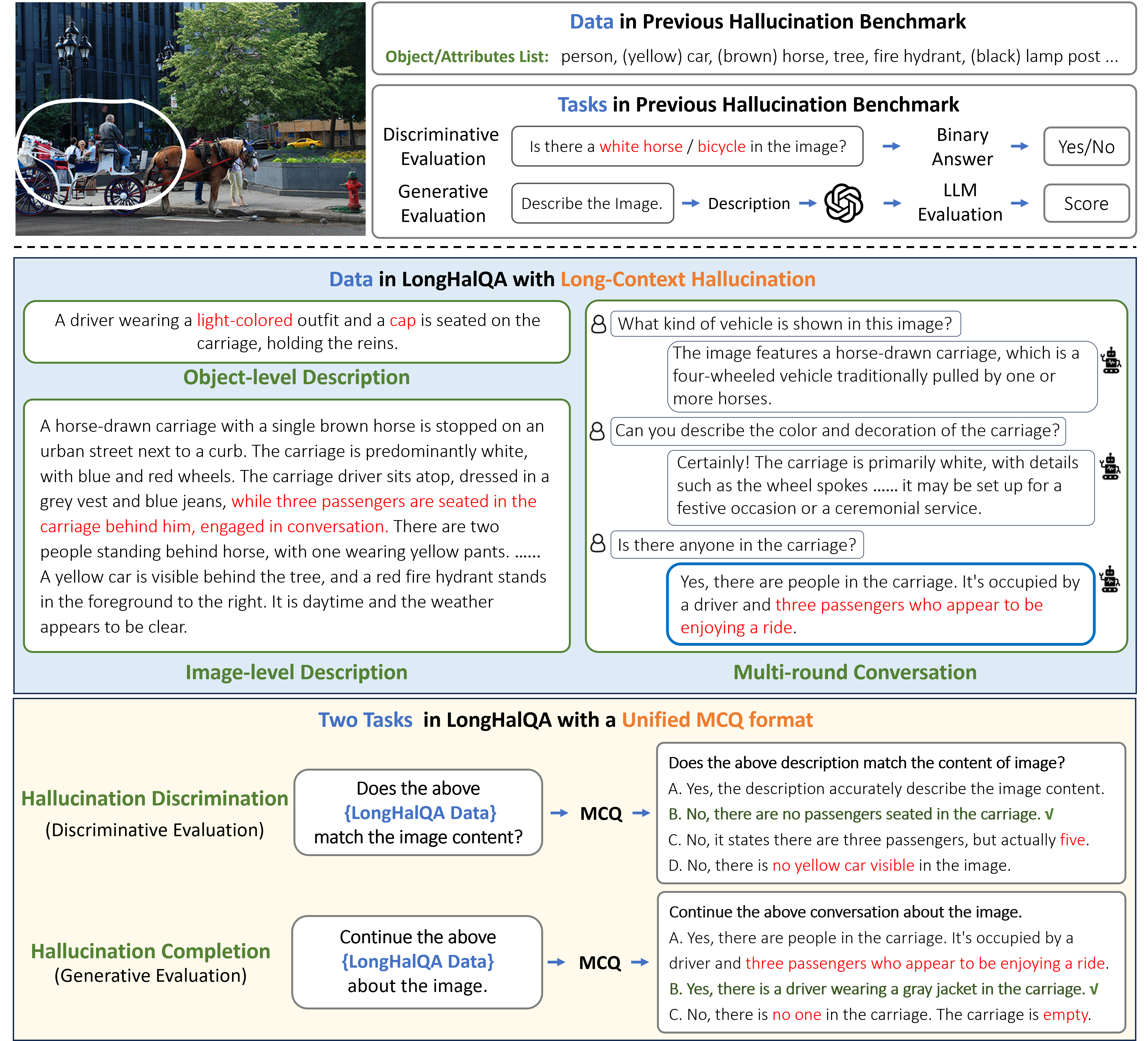}
\caption{LongHalQA is featured with two novel tasks, namely, \textit{Hallucination Discrimination} and \textit{Hallucination Completion}, which unify both discriminative and generative evaluations into the same multiple-choice-question form without requiring costly LLM evaluations. It comprises three types of long-context data, including \textit{Object-level Description}, \textit{Image-level Description}, and \textit{Multi-round Conversation}. Compared with short and simple questions in existing benchmarks like "Is there an \{object\} in the image?", the three types of data are more open-ended, richer in contextual information, and closer to real-world data. White circle in image emphasizes the hallucination of passengers.}
\label{fig:data_format}
\vspace{-20pt}
\end{figure}

Several related benchmarks~\citep{pope,qiu2024valor,jiang2024hal,liu2024phd,lovenia2023negative,wang2023amber,wang2024mitigating} have been proposed to gauge the hallucination level of MLLMs in two representative approaches. The first approach conducts discriminative evaluations, where MLLMs are queried with simple questions about whether some objects exist in the image, as illustrated in the upper part of Fig.~\ref{fig:data_format}. The second approach conducts generative evaluations, which first apply MLLMs to describe the image and then adopt LLM evaluators to examine whether MLLMs generate hallucinatory content. However, most existing benchmarks share several constraints: 1) Most discriminative benchmarks merely require a yes-or-no answer, which is often too simple to tell much on the cause of hallucinations. 2) Discriminative benchmarks usually come with very short questions like "Is there an \{object\} in the image?" which are oversimplified and insufficient for examining hallucination in sophisticated real-world scenarios. 3) Both discriminative and generative benchmarks~\citep{pope,qiu2024valor,kaul2024throne} usually leverage off-the-shelf object annotations to construct questions or detect hallucinatory objects, leading to limited variability~(e.g., a fixed set of 80 object categories for COCO) and biased evaluations toward a small set of objects. 4) Generative benchmarks~\citep{kaul2024throne,jiang2024hal,qiu2024valor,sun2023aligning,liu2023mitigating} generally employ LLMs for hallucination evaluations, but LLMs are computationally intensive and often unstable due to their inherent randomness.

We design LongHalQA, an LLM-free hallucination benchmark that comprises 6K long and context-rich hallucination questions. LongHalQA is built from GPT4V-generated hallucinatory data that is well aligned with various real-world scenarios. It features two multiple-choice-question~(MCQ) tasks, namely, hallucination discrimination and hallucination completion as illustrated at the bottom of Fig.~\ref{fig:data_format}. Specifically, hallucination discrimination requires MLLMs to determine whether the given text contains hallucinations and pick the \textit{right causes} of the hallucinations. Hallucination completion instead transforms generative evaluations into a discriminative task, asking MLLMs to continue the text and pick the right option that does not contain hallucinations. LongHalQA thus unifies discriminative and generative evaluations into the same MCQ form, assessing MLLMs’ understanding of hallucinations and their tendency to generate hallucinations concurrently. LongHalQA queries come in three data formats as illustrated in the middle of Fig.~\ref{fig:data_format}, including object-level descriptions, image-level descriptions, and multi-round conversation, which are much longer and complex, covering a wide range of 12 types of hallucinations. Such long and complex questions allow LongHalQA to gauge the hallucination levels of MLLMs in more practical applications and scenarios. Compared to existing generative benchmarks, we demonstrate that our MCQ hallucination completion task exhibits similar trends to free-form generative evaluation. Additionally, LonghalQA achieves much higher speed in evaluating MLLMs generative hallucination, especially for extremely large models, facilitating the expansion of evaluation samples and the need for fast testing and evolution of MLLMs.

Additionally, we propose LongHallGen, an automated pipeline for \textbf{Long}-context \textbf{Hall}ucination Data \textbf{Gen}eration. LongHallGen is featured with a set of prompt templates for GPT4v that allow generating hallucination data and converting the generated data into multiple-choice-questions automatically. By modifying the prompt sets, LongHallGen can adjust the type of generated hallucinations, content topics, and data formats. We believe that LongHallGen will serve as a strong basis while creating new or expanding existing hallucination datasets for evaluating and training MLLMs in future research.

Based on LongHalQA, we evaluate ten mainstream MLLMs on long-context hallucinations and provide a comprehensive analysis. The evaluations reveal constraints of MLLMs in discerning and explaining hallucination in long texts, as well as in generating hallucinatory content when completing long texts. Additionally, we observe that the Chain-Of-Thought~(COT), a simple but effective hallucination mitigation method~\citep{jiang2024hal,qian2024easy}, is effective for short queries and generative hallucinations but degrades the performance of most MLLMs on long-context hallucinations discrimination in LongHalQA, especially for those with small sizes,  This suggests that COT may be limited by MLLMs' capability on long context processing. We believe that LongHalQA will serve as a basis for mitigating long-context hallucinations in various real-world MLLM tasks.

\section{Related Works}

\noindent\textbf{Hallucination Benchmarks for MLLMs.} Various benchmarks have been proposed to measure the hallucination level of MLLMs, including both discriminative and generative benchmarks. For discriminative benchmarks, POPE~\citep{pope}, CIEM~\citep{ciem}, AMBER~\citep{wang2023amber}, NOPE~\citep{lovenia2023negative}, and MME~\citep{mme} query MLLMs with simple questions about the existence or attributes of specific objects in images. PhD~\citep{liu2024phd} and Hal-Eval~\citep{jiang2024hal} introduce more types of intrinsic hallucinations into evaluations, such as multi-modal conflicting or event hallucinations. Most of these benchmarks feature simple and short questions and seek solely simple binary "yes" or "no" answers. Generative evaluations introduce LLM evaluators to analyze hallucinations in MLLM-generated text. Various generative benchmarks~\citep{jiang2024hal,kaul2024throne,qiu2024valor,wang2023amber} are proposed to improve the scope of evaluated hallucinations and the efficiency and accuracy of LLM evaluators. Our LonghalQA takes a different perspective on long-context hallucinations and encompasses both discriminative and generative evaluations in a unified and efficient MCQ format.

\section{LongHalQA: Long-Context Hallucination Benchmark}

LongHalQA comprises 6485 multiple-choice questions~(MCQ) that cover two tasks with long-context text: hallucination discrimination and hallucination completion. This section presents the task format in Sec.~\ref{sec:task_format}, the data format and distributions in Sec.~\ref{sec:data_format_distribution}, and the evaluation metrics in Sec.~\ref{sec:metrics}.

\subsection{Task Format}
\label{sec:task_format}

In order to comprehensively evaluate the hallucination level of MLLMs, we introduce two tasks, namely, hallucination discrimination and hallucination completion, which conduct discriminative and generative MLLM evaluations, respectively.

\noindent\textbf{Hallucination Discrimination.} For discriminative evaluations, we propose a set of multiple-choice questions to query MLLMs whether object/image descriptions or text responses in a conversation match the contents of images as illustrated in Fig.~\ref{fig:data_format}. Each hallucination question is equipped with multiple answer choices and corresponding explanations. One of the choices starts with "yes," suggesting that the text matches the image contents, while the other three start with "no," followed by explanations. MLLMs are required not only to identify the presence of hallucination but also to understand why hallucination happens and choose the correct explanation. LongHalQA comes with 4346 such image-question samples for the hallucination discrimination task.

\noindent\textbf{Hallucination Completion.} Previous generative benchmarks require MLLMs to generate descriptions for the image and employ LLM evaluators to score descriptions. To obviate the slow generation process and costly LLM evaluations, we transform generative evaluations into MCQ format, as shown in Fig.~\ref{fig:data_format}. Specifically, we provide an image and a related incomplete description or conversation and ask MLLMs to continue the text. Four answer choices of possible completing sentences are provided, with one correct choice and three hallucinatory choices. Compared to generative benchmarks based on LLM evaluators, the format of generative MCQ significantly reduces evaluation costs and allows for more detailed annotation and analysis of hallucination data. LongHalQA comes with 2149 samples for the hallucination completion task.

The MCQ hallucination completion task simulates the sampling process in MLLM inference, where MLLMs first generate several potential outputs and then select the most appropriate one free from hallucinations. Furthermore, the hallucination completion task can be adapted into a free-form continuation task, where MLLMs are prompted to freely continue long-context data. Our experiments demonstrate that the MCQ format of the hallucination completion task and the free-form generation format yield similar trends in evaluating generative hallucinations of MLLMs.

\begin{table}[t]
\footnotesize
    \centering
    \caption{Statistics of 12 types of hallucinations in LongHalQA. ``Object'', ``Description'', and ``Conversation'' denote the data formats of object-level descriptions, image-level descriptions, and multi-round conversations, respectively. We use both hallucinatory and non-hallucinatory data to construct the discrimination task, while only hallucinatory data for the completion task.}
    \begin{tabular}{lcccc}
        \toprule
        Hallucination Types & Object & Description & Conversation & Total \\
        \midrule
        H1 \qquad (Non) Existent Objects & 234 & 261 & 323 & 818 \\
        H2 \qquad Object Attributes & 89 & 130 & 175 & 394 \\
        H3 \qquad Object Color & 122 & 90 & 86 & 298 \\
        H4 \qquad Object States & 50 & 67 & 84 & 201 \\
        H5 \qquad Number of Objects & 80 & 92 & 134 & 306 \\
        H6 \qquad Object Locations & 45 & 76 & 86 & 207 \\
        H7 \qquad Object Relationships & 49 & 54 & 49 & 152 \\
        H8 \qquad Text / Sign Meaning & 27 & 61 & 91 & 179 \\
        H9 \qquad Environment Description & 10 & 13 & 31 & 54 \\
        H10 \enspace \quad Background Description & 13 & 14 & 21 & 48 \\
        H11 \enspace \quad Time   & 2 & 7 & 5 & 14\\
        H12 \enspace \quad Weather & 2 & 5 & 13 & 20\\
        \midrule
        Hallucinatory Data & 723~(52.7\%) & 869~(63.3\%) & 1098~(68.5\%) & 2690~(61.9\%)\\
        Non-hallucinatory Data & 647~(47.2\%) & 503~(36.7\%) & 506~(31.5\%) & 1656~(38.1\%) \\
        \midrule
        \textbf{Data for Hallucination Discrimination} & 1370 & 1372 & 1604 & 4346 \\
        \textbf{Data for Hallucination Completion} & - & 869 & 1270 & 2139 \\
        \midrule
        Total & 1370 & 2241 & 2874 & 6485 \\
        \bottomrule
    \end{tabular}
    \label{tab:hall_types}
\end{table}

\subsection{Data Format and Distribution}
\label{sec:data_format_distribution}
This section presents the format of long-context hallucination data from two aspects, namely, data formats and types of hallucinations, more details to be elaborated in the ensuing subsections.

\noindent\textbf{Data Formats.}
As shown in Tab.~\ref{tab:hall_types}, LongHalQA consists of three formats of image-text hallucinatory data, including 1370 \textit{Object-level Description}, 1372 \textit{Image-level Description}, and 1604 \textit{Multi-round Conversation}. Specifically, \textit{Object-level Description} describes a specific object in the image, such as its attributes, states, or relations with other objects. \textit{Image-level Description} covers the main contents and more details of an image, such as objects, background, weather, etc, in one paragraph. For the \textit{Multi-round Conversation}, we simulate a human user who communicates with an assistant, querying the image content. The three types of data formats are highly compatible with the actual application scenarios of MLLMs and thus can better simulate real hallucination situations.

\begin{figure}[t]
\centering
\includegraphics[width=0.95\linewidth]{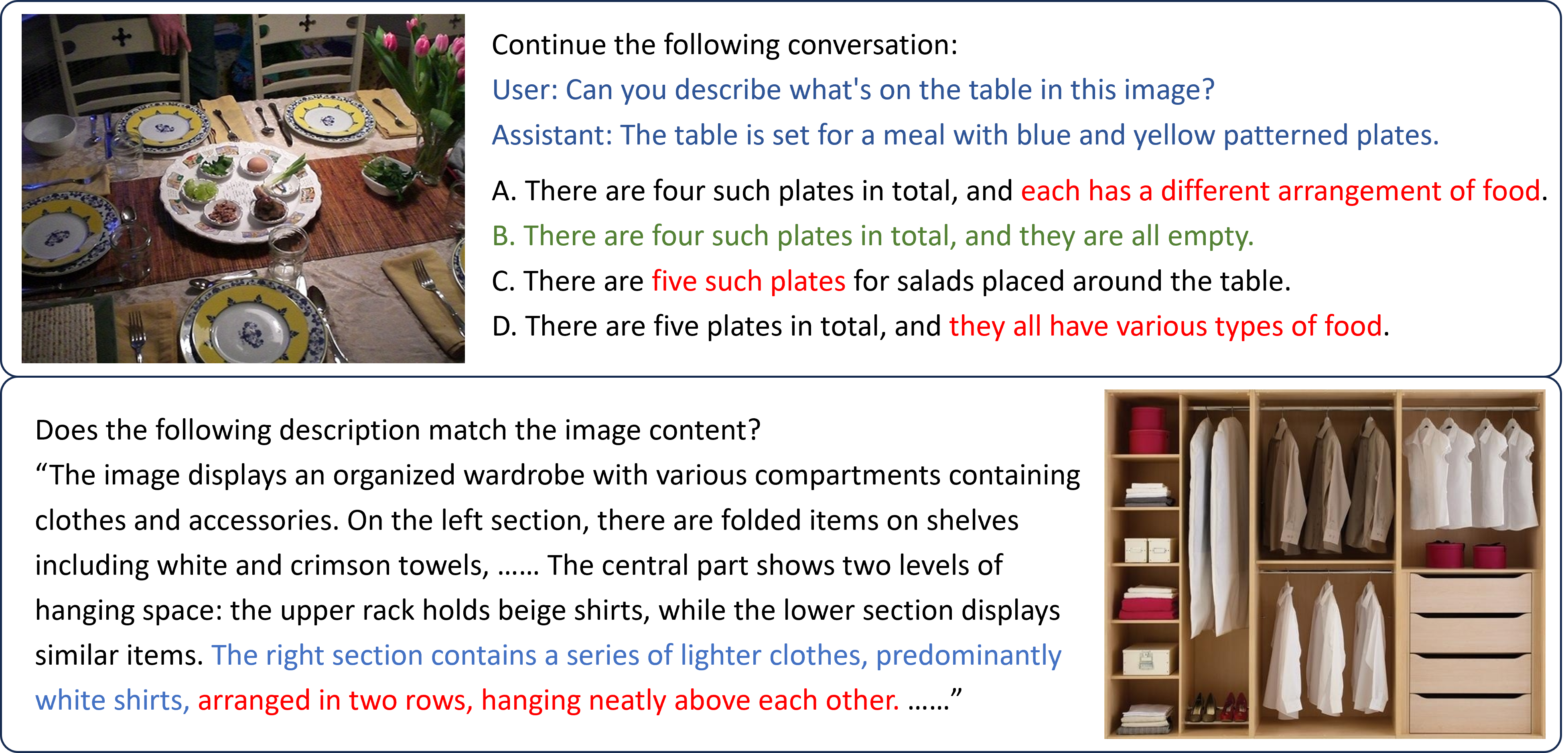}
\caption{LongHalQA involves complex hallucination annotations involving logic and textual consistency, which are closer to hallucinations in real-world MLLM application scenarios.}
\label{fig:hall_example}
\vspace{-10pt}
\end{figure}

\noindent\textbf{Types of Hallucination.}
As shown in Tab.~\ref{tab:hall_types}, We analyze the description texts of objects/images and categorize the wrong descriptions into twelve types of hallucinations. The detailed definitions and visualization of each type of hallucination are provided in Appendix A. Compared to existing benchmarks that focus on the existence and attributes of objects, LongHalQA contains a much broader collection of hallucinations with detailed annotations. 

In addition, LongHalQA includes hallucination samples involving logic and contextual consistency, which are untouched in most existing benchmarks. Such complex hallucinations are often observed with contextually inconsistent descriptions such as 'four such plates' or 'five plates' as illustrated in the upper part of Fig.~\ref{fig:hall_example}, or incorrectly mixed descriptions such as 'two rows of shirts in the central part' vs 'one row of shirts in the right part' as illustrated in the lower part of Fig.~\ref{fig:hall_example}. We foresee that LongHalQA will inspire more in-depth studies of MLLMs regarding such complex hallucinations.

\noindent\textbf{Complexity and Length of Text in LongHalQA.}
We derive certain statistics of LongHalQA data to verify their quality, including the length and the number of object nouns appearing in the text data. The study shows LongHalQA has an average of 14/130 and 189 words, respectively, for object/image-level descriptions and multi-round conversations, clearly longer and more informative than descriptions of around 80 words in existing generative benchmarks~\citep{jiang2024hal,kaul2024throne}. In addition, LongHalQA contains approximately 4K object names, indicating more informative data compared with existing benchmarks~\citep{pope,ciem} with fixed annotations (e.g., 80 object names within COCO dataset).

\subsection{Evaluation Methods and Metrics}
\label{sec:metrics}

We adopt both binary and multiple-choice settings for the Hallucination Discriminiation task, and multiple-choice setting for Hallucination Completion task. For binary answers, we use Accuracy, Precision, and "Yes" ratios as metrics following previous practices~\citep{pope,jiang2024hal}. For multiple-choice setting, we adopt (mc-)accuracy~\citep{MMBench} as the evaluation metric, which requires MLLMs to generate the letter (e.g., A, B, C, or D) of the answer option.
We randomly shuffle the order of the four options for each MCQ to reduce the impact of option order.

\section{LongHallGen: Automated \textbf{Long}-Context \textbf{Hall}ucination data \textbf{Gen}eration}

Given the lack of long-context image-text hallucination data in the broad area of vision language understanding, we dig deep into the proposed LongHallQA and distill LongHallGen, a generic pipeline that aims to facilitate the construction of long-context hallucination benchmarks or datasets in various multimodal tasks, more detailed processes to be elaborated in the ensuing subsections.

\noindent\textbf{Image Collection and Filtering.} To generate informative hallucinatory data, the first step involves selecting images that contain rich content and discarding those with overly simple scenes or rare objects. This can be simply achieved by leveraging off-the-shelf image understanding techniques. We adopt images from the validation set of VisualGenome~\citep{krishna2017visual} and Objects365~\citep{objects365} to avoid images being used for training MLLMs, and then analyze and filter them based on dataset annotations and GroundingDINO~\citep{groundingdino}. This straightly leads to 1200 images with complex contents for further hallucinatory data generation.

\noindent\textbf{Positive Data Generation.} With selected images, long-context data can be generated by leveraging state-of-the-art MLLMs. This process involves designing certain prompts according to specific tasks and domains. We adopt GPT4V to generate long-context texts. Unified prompt templates are designed to allow adjusting the format and scope of generated texts, as illustrated in Appendix B. Note current MLLMs, even GPT4V, suffer from severe hallucinations while generating long-context data. A hallucination check process is required to ensure the quality of the generated data.

\noindent\textbf{Hallucination Check.} The MLLM-generated data are then examined comprehensively for detecting any inherent hallucinations. The examination can be achieved in three steps. First, the generated data undergo a per-sentence self-check by GPT4V, where each piece of data is checked twice to reduce randomness. Second, names of objects present in the data are extracted, and certain image understanding tools such as GroundingDINO~\citep{groundingdino} are then conducted with detection results feeding to GPT4V for further checking. Finally, the summarized analysis are revised manually, and the revised analysis can serve for the generation of hallucination-explanation data pairs.

Though MLLMs such as GPT-4V can generate lengthy descriptions, they tend to generate numerous hallucinations as well. Take LongHallQA as an example. Among its generated descriptions for the 500 images on image-level descriptions, 394 descriptions (78.8\%) contain at least one hallucination. The ratio goes up to 82.4\% for the generated conversations. One major cause of the high hallucination rate is due to the increased length of the generated descriptions. In addition, most selected images in LongHalQA have complex scenes, further boosting the possibility of descriptive hallucinations. Nevertheless, such data realistically simulate the hallucinations in actual applications of MLLMs.

\noindent\textbf{Hallucination-Explanation Pair Generation.}
Two different prompts are formulated to construct hallucination-explanation~(HE) data pairs with the checking outcome. For the data without hallucinations detected, GPT4V is prompted to modify the data to produce a misleading error within the range of hallucination types suggested in the prompt. For the data containing hallucinations, GPT-4V is prompted to modify the data to contain only one error to form HE pairs. The generated HE pairs are then adopted to construct MCQs for MLLM evaluations.

\begin{table}[t]
    \footnotesize
    \centering
    \caption{Evaluations of MLLMs on LongHalQA with Hallucination Discrimination and Hallucination Completion tasks with binary answer accuracy~(`bi-Acc') and multi-choice accuracy ('mc-ACC').}
    \begin{tabular}{lcccc}
         \toprule
         & \multicolumn{2}{c}{Hallucination Discrimination} & Hallucination Completion & \multirow{2}{*}{Average} \\
         Model & bi-Acc. & mc-ACC. & mc-ACC. & \\
         \midrule 
        MiniCPM-V2-2B & 44.15 & 40.80 & 46.25 & 43.73 \\
        Qwen2-VL-2B & 48.21 & 38.76 & 50.36 & 45.78 \\
        \midrule
        Fuyu-8B  & 43.31 & 23.86 & 23.67 & 30.28 \\
        LLaVA-1.5-7B & 38.52 & 35.04 & 36.08 & 36.55 \\ 
        LLaVA-1.5-13B & 41.83 & 43.60 & 37.58 & 41.00 \\
        LLaVA-1.6-7B & 44.13 & 45.56 & 43.40 & 44.36\\
        Qwen-VL-Chat & 43.21 & 37.03 & 36.57 & 38.94\\
        \midrule
        LLaVA-1.6-34B & 46.99 & \textbf{57.40} & 56.03 & 53.47\\
        Qwen2-VL-72B & 50.36 & 54.78 & \textbf{61.50} & \textbf{55.55} \\
        \midrule
        GPT4o & \textbf{52.80} & 47.63 & 56.15 & 52.19 \\
         \bottomrule
    \end{tabular}
    \label{tab:result_all}
\end{table}

\noindent\textbf{Question and Answer Generation.}
With the generated HE pairs, MLLMs such as GPT-4V can be employed to generate questions for tasks such as hallucination discrimination and hallucination completion in LongHalQA. For discriminative tasks, questions like 'Does the following \{Hallucination Data\} match the image?' can be formulated to prompt MLLMs to generate four candidate options with explanations. For completion tasks with questions like 'Complete the following  \{Hallucination Data\}.', MLLMs are employed to construct a completion task by providing prefix text from HE pairs and options of candidate sentences for completion.

LongHallGen exploits MLLMs for most processes in generating long-context hallucination data, except the hallucination checking that involves optional human verification. We expect LongHallGen to serve as a basis for constructing more long-context hallucination data for training and evaluating MLLMs, thereby enhancing their capability and reliability in complex application scenarios.

\begin{table}[t]
    \footnotesize
    \centering
    \caption{Experiments on LongHalQA for \textbf{Hallucination Discrimination} task with binary answers.``Acc.'', ``Pre.'', and ``YR'' denote accuracy, precision, and Yes ratio, respectively.}
    \begin{tabular}{lccccccccc}
    \toprule
        & \multicolumn{3}{c}{Object-level Description} & \multicolumn{3}{c}{Image-level Description} & \multicolumn{3}{c}{Multi-round Conversation.} \\
        Model & Acc. & Pre. & YR & Acc. & Pre. & YR & Acc. & Pre. & YR \\
        \midrule 
        MiniCPM-V2-2B & 59.71 & 54.67 & 74.23 & 36.66 & 36.64 & 99.85 & 36.10 & 31.91 & 89.28 \\
        Qwen2-VL-2B & 64.31 & 58.01 & 71.97 & 36.88 & 36.74 & 99.78 & 43.45 & 32.05 & 69.45 \\
        \midrule
        Fuyu-8B & 50.29 & 48.52 & 83.65 & \textbf{43.29} & 37.28 & 78.79 & 36.35 & 30.72 & 82.79 \\
        LLaVA-1.5-7B & 45.18 & 45.99 & 94.60 & 36.59 & 36.62 & 99.93 & 33.79 & 47.53 & 94.58 \\ 
        LLaVA-1.5-13B & 52.70 & 49.96 & 84.16 & 36.95 & 36.95 & 99.71 & 35.85 & 31.93 & 90.02 \\
        LLaVA-1.6-7B & 60.51 & 55.31 & 72.85 & 37.10 & 36.82 & 99.56 & 34.79 & 31.90 & 92.83 \\
        Qwen-VL-Chat & 58.69 & 53.71 & 79.64 & 36.66 & 36.66 & 100.0 & 34.29 & 31.78 & 93.58 \\
        \midrule
        LLaVA-1.6-34B & 68.61 & 61.65 & 67.96 & 38.26 & 37.22 & 98.10 & 34.10 & 32.16 & 96.13 \\
        Qwen2-VL-72B & 71.60 & 64.46 & 65.11 & 40.08 & 37.87 & 95.84 & 39.40 & 33.59 & 88.34 \\
        \midrule
        GPT4o & \textbf{73.94} & 68.31 & 57.81 & 37.92 & 37.13 & 97.03 & \textbf{46.32} & 35.71 & 77.74 \\
         \bottomrule
    \end{tabular}
    \label{tab:result_binary}
\end{table}

\begin{minipage}[t!]{0.45\textwidth}
\footnotesize
\centering
\makeatletter\def\@captype{table}
\caption{Experiments on \textbf{Hallucination Discrimination} under multi-choice settings. ``Desc'' indicates description.}
\vspace{5pt}
\label{tab:discrim_mcq}
\begin{tabular}{lcc}
    \toprule
    Accuracy & Image Desc. &  Conversation \\
    \midrule
    MiniCPM-V2-2B & 39.65 & 41.96 \\
    Qwen2-VL-2B & 41.55 & 35.97 \\
    \midrule
    Fuyu-8b & 23.47 & 24.25 \\
    LLaVA-1.5-7B & 37.17 & 32.92 \\ 
    LLaVA-1.5-13B & 45.99 & 41.21 \\
    LLaVA-1.6-7B & 49.42 & 41.71 \\
    Qwen-VL-Chat & 37.97 & 36.10 \\
    \midrule
    LLaVA 1.6-34B & \textbf{60.93} & 53.86 \\
    Qwen2-VL-72B & 53.57 & \textbf{55.98} \\
    \midrule
    GPT-4o & 46.57 & 48.69 \\
    \bottomrule
\end{tabular}
\end{minipage}\hspace{20pt}
\begin{minipage}[t!]{0.45\textwidth}
\footnotesize
\centering
\makeatletter\def\@captype{table}
\caption{Experiments on \textbf{Hallucination Completion} under multi-choice settings.``Desc'' indicates discription.}
\vspace{5pt}
\label{tab:completion_mcq}
\begin{tabular}{lcc}
    \toprule
    Accuracy & Image Desc. &  Conversation \\
    \midrule
    MiniCPM-V2-2B & 44.07 & 48.43 \\
    Qwen2-VL-2B & 47.18 & 53.54 \\
    \midrule
    Fuyu-8b & 23.25 & 24.09 \\
    LLaVA-1.5-7B & 32.80 & 39.37 \\ 
    LLaVA-1.5-13B & 31.53 & 43.62 \\
    LLaVA-1.6-7B & 39.47 & 47.32  \\
    Qwen-VL-Chat & 33.14 & 40.00 \\
    \midrule
    LLaVA-1.6-34B & 53.16 & 58.90 \\
    Qwen2-VL-72B & \textbf{59.38} & \textbf{63.62} \\
    \midrule
    GPT-4o & 50.97 & 61.33 \\
    \bottomrule
\end{tabular}
\end{minipage}

\section{Experiments}

\subsection{Overall Experiments}
We adopt LMMs-Eval~\citep{lmms_eval2024} to employ LongHalQA to gauge the hallucination level of MLLMs. The evaluations are performed over nine widely adopted open-source MLLMs, including MiniCPM-V2~\citep{hu2024minicpm}, Qwen series~\citep{Qwen-VL,wang2024qwen2}, Fuyu~\citep{fuyu8b2023}, LLaVA series~\citep{llava15,liu2024llavanext}, and the closed-source GPT-4o, covering MLLMs' sizes from 2B to 72B and larger. We present the overall experiments in Tab.~\ref{tab:result_all}. We use GPTQ-Int8 quantization for Qwen2-VL-72B due to memory constraints. Notably, Qwen2-VL-72B achieves the best accuracy on average for hallucination completion tasks, demonstrating its superior ability to identify hallucinated information and produce reliable content. Such performance advantage may be attributed to the superior capabilities of LLM and their proposed naive dynamic resolution mechanism. Following Qwen2-VL-72B are LLaVA-v1.6-34B and GPT-4o. This performance advantage suggests GPT's potential capability of self-correction for hallucinations, given that the LongHalQA is primarily based on hallucination data from GPT. Next, smaller models like MiniCPM-V2, Qwen2-VL-2B, and and LLaVA 1.6-7B also achieved excellent results, surpassing many larger models. It is worth noting that both MiniCPM-V2 and Qwen2-VL-2B adopts reinforcement learning to mitigate hallucinations, indicating that this is an effective method for improving the reliability of MLLMs. One common feature of these leading MLLMs is that they all support high-resolution images, suggesting that resolution plays a significant role in alleviating hallucinations.

\subsection{Experiments on Hallucination Discrimination}

\noindent\textbf{Binary-Answer Setting.}
Tab.~\ref{tab:result_binary} shows experiments on the hallucination discrimination task under the binary answer setting. GPT-4o performs the best among all MLLMs, particularly for the multi-round conversations, with an accuracy gain of 9.5\% over other MLLMs. Fuyu-8b shows superior capabilities in identifying hallucinations in long text and achieves the best accuracy among all open-source MLLMs, scoring 45.0\% for image-level descriptions and 36.8\% for multi-round conversations. We observe that most MLLMs produce a high yes ratio of over 70\%, even 99\% for image-level description, largely deviating from the ratio of non-hallucinatory data with answers 'yes' in LongHalQA (38.1\%). Moreover, Qwen-VL-Chat, MiniCPM-V2-2B, and LLaVA series exhibit unbalanced capabilities in handling text of varying lengths. For object-level descriptions, LLaVA1.6-7B achieves an accuracy of 60.6\%, but this drops to 37.1\% and 34.7\% for image-level descriptions and conversations that are about ten times longer. The lower accuracy, coupled with a significantly high Yes ratio, demonstrates the constraints of existing MLLMs in recognizing hallucinations in long contexts.

\begin{table}[h]
    \footnotesize
    \centering
    \caption{Experiments on different types of hallucinations on discrimination task with multiple-choice setting. The indexes of hallucination types are consistent with Tab.~\ref{tab:hall_types}.}
    \begin{tabular}{lcccccccccccc}
         \toprule
          & H1 & H2 & H3 & H4 & H5 & H6 & H7 & H8 & H9 & H10 & H11 & H12 \\
         \midrule 
        MiniCPM-V2-2B & 20.1 & 19.5 & 16.5 & 18.3 & 22.7 & 20.7 & 21.8 & 16.7 & 16.7 & 16.7 & 33.3 & 17.6\\
        Qwen-VL-Chat & 12.4 & 10.1 & 10.0 & 9.8 & 10.0 & 12.9 & 11.9 & 8.3 & 16.7 & 6.7 & 33.3 & 11.7 \\
        Fuyu-8B & 27.9 & 26.9 & 30.1 & 25.3 & 26.3 & 22.0 & 20.8 & 27.8 & 26.2 & 30.0 & 40.0 & 29.4 \\
        LLaVA-1.5-7B & 9.0 & 9.1 & 8.8 & 4.2 & 10.0 & 8.8 & 12.9 & 6.9 & 19.0 & 10.0 & 20.0 & 5.9  \\ 
        LLaVA-1.5-13B & 22.0 & 22.9 & 26.6 & 23.2 & 26.4 & 30.8 & 26.7 & 26.4 & 35.7 & 26.7 & 26.7 & 35.3 \\
        LLaVA-1.6-7B & 32.8 & 40.4 & 40.2 & 30.9 & 30.8 & 43.4 & 43.5 & 38.8 & 47.6 & 20.0 & 20.0 & 47.0 \\
         \bottomrule
    \end{tabular}
    \label{tab:result_per_hall_type}
\end{table}

\noindent\textbf{Multiple-choice Setting.}
Tab.~\ref{tab:discrim_mcq} shows experiments on the hallucination discrimination task under the MCQ setting. LLaVA-1.6-34B and Qwen2-VL-72B achieve the highest accuracy in description and conversation data formats, respectively, followed closely by GPT-4o. Notably, most MLLMs achieve much higher accuracy in both image description and conversation formats compared to the accuracy under binary settings. This is likely because answer choices include detailed explanations for the involved hallucinations, giving the model a clearer understanding and aiding in selecting the correct option. However, the much lower ranking-based accuracy suggests that MLLMs struggle to correctly discern hallucinations and provide accurate reasons when they cannot directly access all options for reference, consistent with the low accuracy observed in binary settings where explanations are also inaccessible. Interestingly, Fuyu-8B achieves the highest ranking-based accuracy, even surpassing its generation-based accuracy, which may be attributed to its unique decoder-only structure.

\subsection{Experiments on Hallucination Completion}
Tab.~\ref{tab:completion_mcq} shows experiments on the hallucination completion task. Among open-sourced MLLMs, Qwen2-VL-72B achieves the best performance for both image description and multi-round conversation. Two small models, MiniCPM-V2 and Qwen2-VL-2B, also achieved excellent results in this task, ranking just behind three much larger models with the least size of 34B. This further reflects the significance of high-resolution representation and multi-modal RLHF~\citep{2023rlhf-v}, which aligns vision and language for trustworthy behavior against object hallucinations in training. Other MLLMs achieve similar ranking performance as those in the hallucination discrimination task, with LLaVA 1.6-7B prevailing, followed by Qwen-VL-Chat and LLaVA 1.5-13B.

\subsection{Analysis of hallucination types.}
We conduct a detailed analysis of different types of hallucinations in the discrimination task as shown in Tab.~\ref{tab:result_per_hall_type}. We find that most MLLMs exhibit relatively higher accuracy in hallucinations of object existence~(H1), attributes~(H2), and colors~(H3), which are relatively simple to discern because they can be directly observed from images and rely less on detailed comprehension of image content. From the perspective of MLLMs, MiniCPM-V2-2B and Qwen-VL-Chat show balanced strength across different types of hallucinations. Fuyu-8B is competitive across multiple types, but struggles with object location~(H6) and relationships~(H7). LLaVA1.6-7B outperforms other MLLMs on most types of hallucination, especially for object location~(H6) and Text/Sign semantic meanings~(H8).

\subsection{Prompt Analysis}
We additionally examine the Chain-Of-Thought~(COT) on LongHalQA, which has been verified in previous benchmarks~\citep{qian2024easy,jiang2024hal} for mitigating hallucinations. We modify the prompt to guide MLLMs to think step by step, and provide more instructions, such as possible types of hallucinations and suggestions for per-sentence verification~(Refer to Appendix C for details). As shown in Tab.~\ref{tab:result_cot}, apart from GPT-4o, most MLLMs experience a drop in performance across different tasks when using Chain of Thought (COT)—notably, the larger the language model, the smaller the performance drop. Qwen2-VL-72B only experienced a drop in the multiple-choice task for hallucination Discrimination, with an average increase of 2.18 accuracy. We conjecture that this is largely due to the limited capability of MLLMs to interpret long-context data. Besides, the modified prompts improve short query discrimination and hallucination completion the most but have little effect on discriminate hallucinations in long texts.

\begin{table}[t]
    \footnotesize
    \centering
    \caption{Experiments on LongHalQA with the modified prompt. `bi-ACC' and `mc-ACC' denote binary answer and multiple-choice accuracies. ``Object'' and ``Long'' indicate data formats of object-level descriptions and long context data of image-level descriptions and conversations.}
    \begin{tabular}{lcccccc}
         \toprule
         & \multicolumn{3}{c}{Hall. Discrimination} & Hall. Completion & \multirow{2}{*}{Average}\\
         Model & (Object) bi-Acc. & (Long) bi-Acc. & mc-ACC. & mc-ACC & \\
         \midrule 
        MiniCPM-V2 & 63.87\color{red}{(+4.14)} & 36.24~\color{blue}{(-0.14)} & 38.12~\color{blue}{(-2.68)} & 48.13\color{red}{(+1.88)} & 43.90\color{red}{(+0.17)} \\
        Qwen2-VL-2B & 61.61~\color{blue}{(-2.70)} & 38.06~\color{blue}{(-2.10)} & 34.95~\color{blue}{(-3.81)} & 50.75\color{red}{(+0.39)} & 43.87\color{blue}{(-1.91)} \\
        \midrule
        Fuyu-8B & 47.23~\color{blue}{(-3.06)} & 34.16~\color{blue}{(-5.66)} & 23.74~\color{blue}{(-0.12)} & 23.35~\color{blue}{(-0.32)} & 28.55~\color{blue}{(-1.73)}\\
        LLaVA-1.5-7B & 49.64~\color{red}{(+4.46)} & 34.84~\color{blue}{(-0.35)} & 34.67~\color{blue}{(-0.37)} & 38.81~\color{red}{(+2.73)} & 37.75~\color{red}{(+1.2)} \\
        LLaVA-1.5-13B & 54.38\color{red}{(+1.68)} & 36.29~\color{blue}{(-0.11)} & 44.12\color{red}{(+0.52)} & 40.90\color{red}{(+3.32)} & 42.45\color{red}{(+1.45)} \\
        LLaVA-1.6-7B & 60.36~\color{blue}{(-0.15)} & 34.68~\color{blue}{(-1.26)} & 40.68~\color{blue}{(-4.88)} & 44.48\color{red}{(+1.08)} & 42.80~\color{blue}{(-1.56)} \\
        Qwen-VL-Chat & 59.85\color{red}{(+1.16)} & 34.76~\color{blue}{(-0.71)} & 35.38~\color{blue}{(-1.65)} & 39.06\color{red}{(+2.49)} & 39.19\color{red}{(+0.25)} \\
        \midrule
        LLaVA-1.6-34B & 66.64~\color{red}{(+1.97)} & 36.03~\color{blue}{(-0.15)} & 45.56~\color{blue}{(-11.84)} & 57.49\color{red}{(+1.46)} & 49.76~\color{blue}{(-3.71)} \\
        Qwen2-VL-72B & 73.58\color{red}{(+1.98)} & 46.08\color{red}{(+6.34)} & 53.11~\color{blue}{(-1.67)} & 64.82\color{red}{(+3.32)} & 57.73\color{red}{(+2.18)}\\
        \midrule
        GPT-4o & 74.84\color{red}{(+0.90)} & 45.50\color{red}{(+3.38)} &  50.12\color{red}{(+2.49)} & \color{red}{58.97(+2.82)} & 54.79\color{red}{(+2.60)} \\
        \bottomrule
    \end{tabular}
    \label{tab:result_cot}
\end{table}

\section{Comparison with free-generation evaluation}

We further compare LongHalQA with evaluations in free-generation scenarios to examine whether the multiple-choice~(MCQ) format of the hallucination completion task in LongHalOA accurately captures the true generative capabilities of MLLMs. As described in Sec.~\ref{sec:task_format}, the multiple-choice hallucination completion task could be transformed into a similar counterpart in a free-form generation setting.  Specifically, we randomly selected 200 image-text data from image description and multi-round conversation data for the hallucination completion task, respectively, and apply MLLMs to complete the text freely. As shown in Tab.~\ref{tab:free_generate_acc}, the ranking of hallucination levels, using GPT-4 evaluation, is largely consistent with the results from LongHalQA, demonstrating that the proposed MCQ task is able to capture the generative capabilities of MLLMs.

\begin{figure}[t]
\footnotesize
\centering
\includegraphics[width=0.7\linewidth]{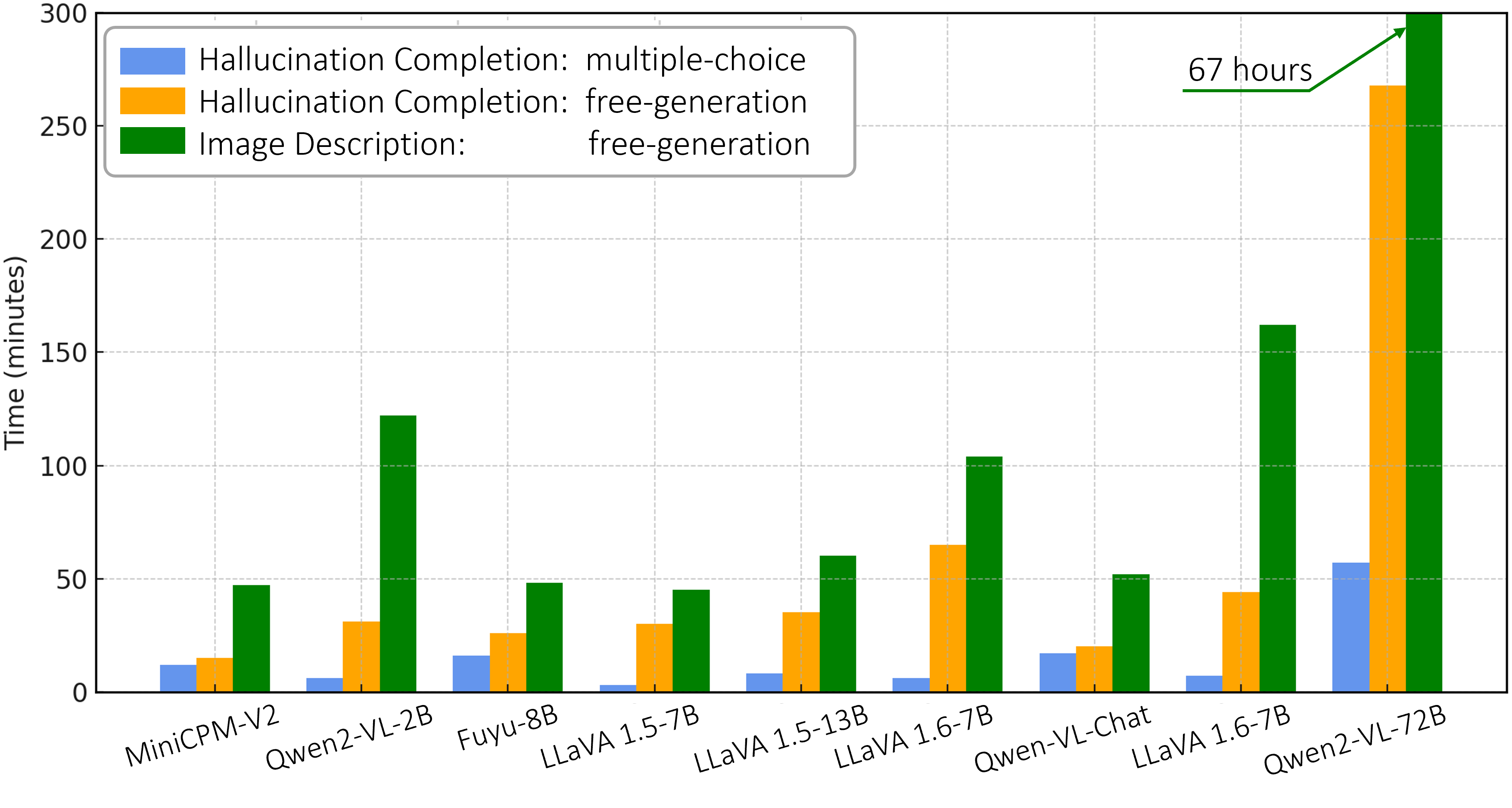}
\caption{Comparison of evaluation times under different settings and MLLMs. Our proposed multiple-choice hallucination completion task is significantly faster than other(existing) setups, especially for large models. We measure the time taken to evaluate 1,000 image-text pairs under three different evaluation settings. Only the time that MLLMs take to generate text is measured without considering the time for evaluations by other LLM evaluators. All MLLMs are tested on one A100 except LLaVA 1.6-34B and Qwen2-VL-72B on an H100.}
\label{fig:eval_time}
\vspace{-20pt}
\end{figure}

Additionally, we compare the efficiency of different approaches in evaluating MLLM's generative hallucination. As shown in Fig~\cref{fig:eval_time}, for some extremely large models, the time required for generating descriptions from scratch may be excessively long and impractical. The multiple-choice format of Longhalqa, while preserving evaluation effectiveness, significantly improves evaluation efficiency and facilitates future expansion of evaluation data in both scale and diversity.

\begin{table}[t]
\footnotesize
\centering
\caption{Comparison of Multi-Choice~(mc-ACC) and Free-Generation~(gen-ACC) settings for \textbf{Hallucination Completion}. Under the free-generation setting, MLLMs are provided with preceding contexts and images and are prompted to freely continue the pretext to assess their generative hallucinations.}
\label{tab:free_generate_acc}
\begin{tabular}{lcccc}
    \toprule
    Accuracy & mc-Acc. & Ranking &  gen-Acc. & Ranking \\
    \midrule
    MiniCPM-V2 & 46.25 & 4 & 54.00 & 5 \\
    Qwen2-VL-2B & 50.36 & 3 & 55.25 & 4 \\
    \midrule
    Fuyu-8b & 23.67 & 9 & 11.50 & 9 \\
    LLaVA 1.5-7B & 36.08 & 8 & 52.50 & 7 \\ 
    LLaVA 1.6-7B & 43.40 & 5 & 59.50 & 3 \\
    LLaVA 1.5-13B & 37.58 & 6 & 53.50 & 6 \\
    Qwen-VL-Chat & 36.57 & 7 & 50.50 & 8 \\
    \midrule
    LLaVA 1.6-34B & 56.03 & 2 & 63.00 & 2 \\
    Qwen2-VL-72B & 61.50 & 1 & 65.75 & 1 \\
    \bottomrule
\end{tabular}
\end{table}

\section{Conclusion}
This paper presents LongHalQA, a novel benchmark with long-context data for evaluating MLLMs' level of hallucinations in more practical scenarios. LongHalQA consists of 6.4k question-answer pairs with long-context data covering 12 types of hallucination. It features two multiple-choice tasks: hallucination recognition and hallucination completion, implementing both discriminative and generative hallucination evaluation in one unified format. It also offers additional assessments of the causes of hallucinations without involving LLM evaluators as in existing benchmarks. We also propose an automated pipeline for generating long-context hallucination data. Extensive tests reveal the constraints of existing MLLMs in handling long-context hallucinations, showing the necessity for more research on robust MLLMs with respect to long-context hallucinations.

\bibliography{iclr2025_conference}
\bibliographystyle{iclr2025_conference}

\appendix
\section{Appendix}
\section{Hallucination Types in LongHalQA}

\begin{figure}[t]
\centering
\includegraphics[width=0.99\linewidth]{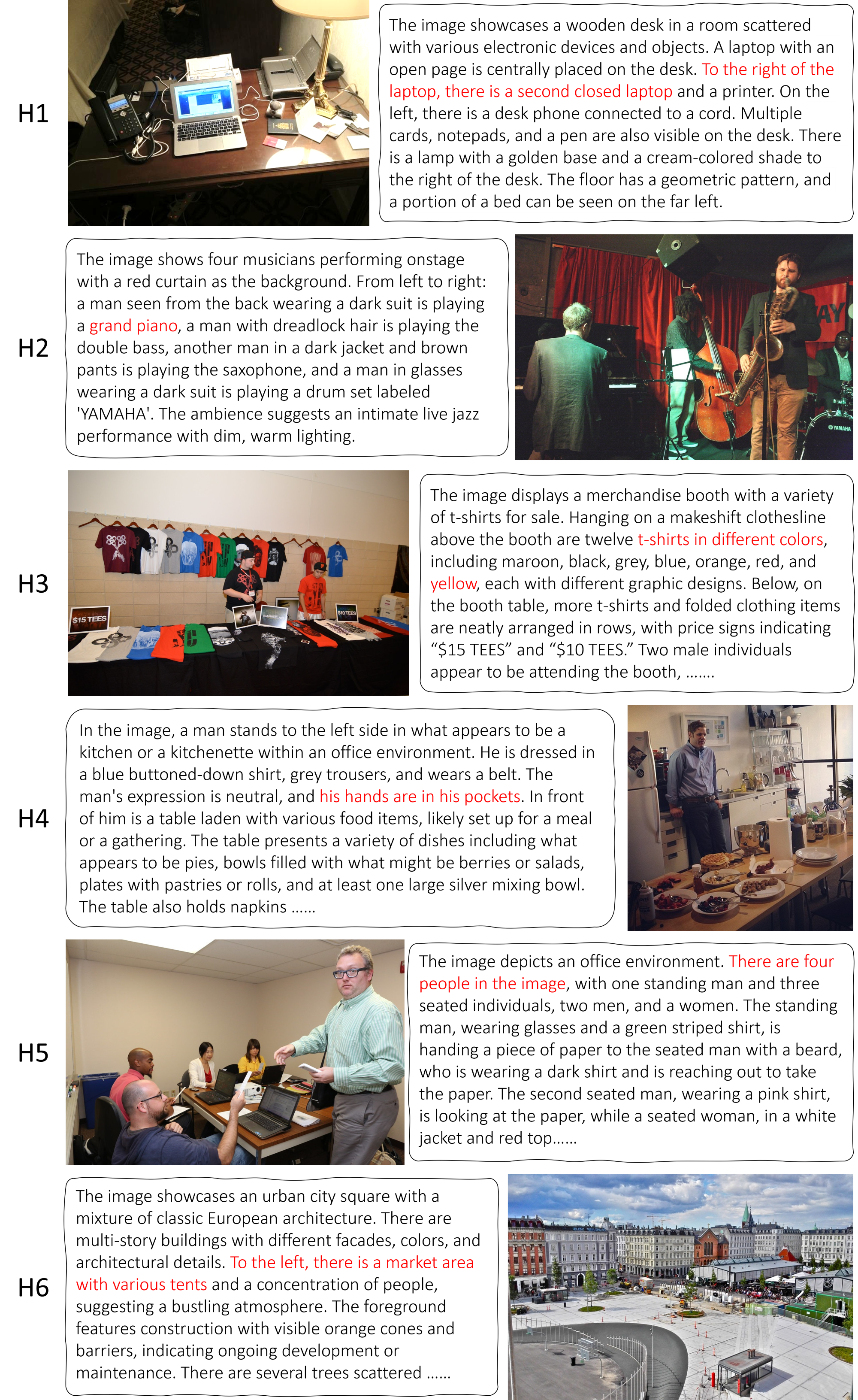}
\caption{Visualizations of hallucination types from H1 to H6.}
\label{fig:hall_visual_1_6}
\end{figure}

\begin{figure}[t]
\centering
\includegraphics[width=0.99\linewidth]{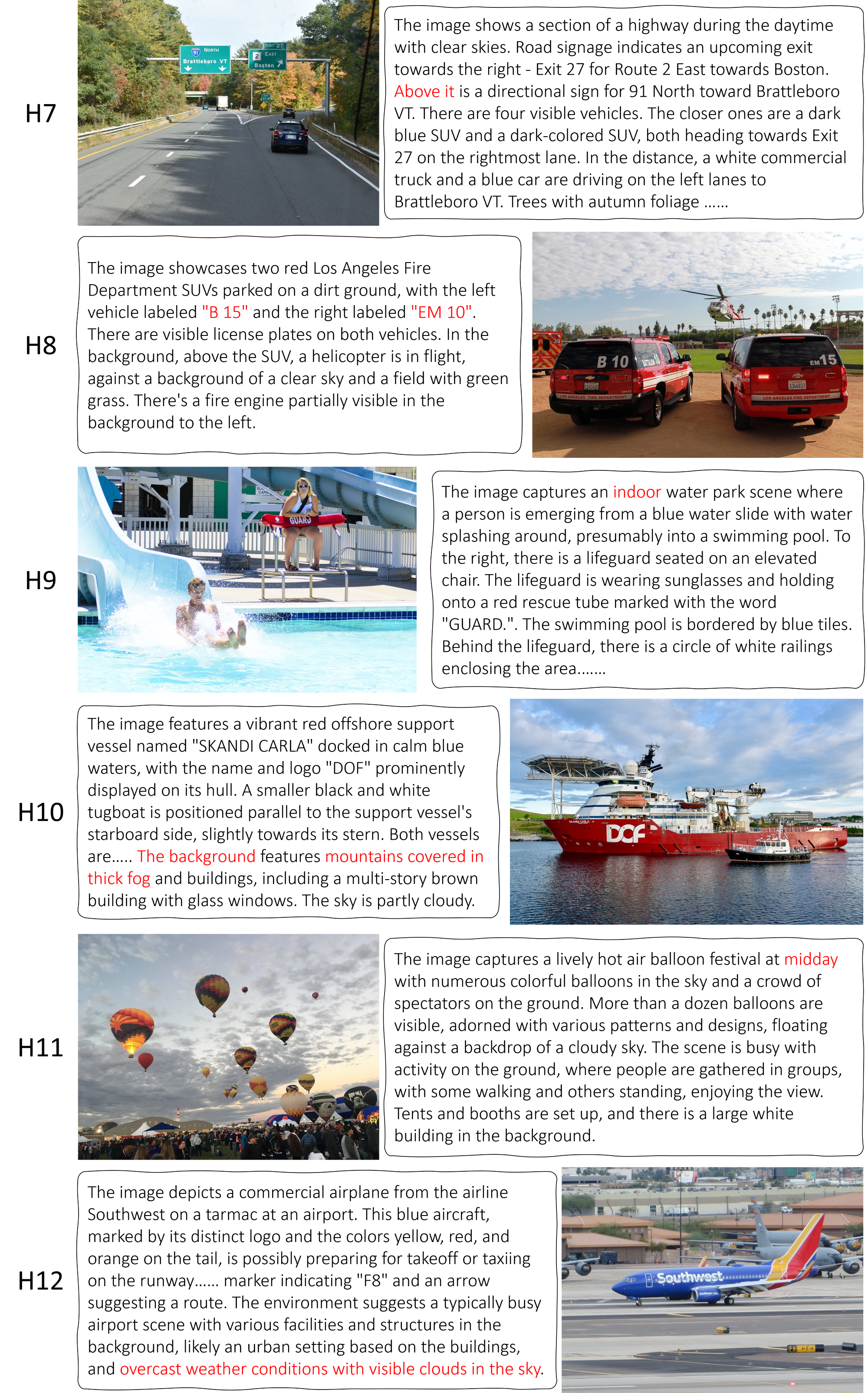}
\caption{Visualizations of hallucination types from H7 to H12.}
\label{fig:hall_visual_7_12}
\end{figure}

We present the definitions of different types of hallucinations as follows. Some examples are shown in Fig.~\ref{fig:hall_visual_1_6} and Fig.~\ref{fig:hall_visual_7_12}.

\begin{itemize}
    \item[H1] \textbf{(Non) Existent Objects.} The described objects do not exist in the given image.
    \item[H2] \textbf{Object Attributes.} The appearances~(shape, pattern, etc.), types, or other attributes of objects are incorrectly described. 
    \item[H3] \textbf{Object Color.} The colors of objects are incorrectly described. (Due to the high number of hallucinations of object color descriptions, we list "object color" separately from object attributes.)
    \item[H4] \textbf{Object States.} The states of objects, such as movement, orientation, the actions of the person, etc, are incorrectly described.
    \item[H5] \textbf{Number of Objects.} The number of objects is incorrectly stated. 
    \item[H6] \textbf{Object Locations.} The locations of objects in the image are incorrectly described.
    \item[H7] \textbf{Object Relationships.} The relations or relative positions between two or multiple objects are incorrectly described.
    \item[H8] \textbf{Text/Sign Meaning.} The text in the image is wrongly discerned, or the meanings of signs, such as street signs, advertisements, price tags, etc, are incorrectly described.
    \item[H9] \textbf{Environment Description.} Wrong descriptions or adjectives of the environment or location, for example, indoors, outdoors, rural, urban, bookstore, food market, etc.
    \item[H10] \textbf{Background Description.} The descriptions of activities, scenes, objects, etc., in the background of the image are hallucinatory. Examples of descriptions include the presence of mountains, buildings, skyscrapers people in the background, or the description of no visible people in the background.
    \item[H11] \textbf{Time.} The incorrect description of the time of day, night, etc., in the image, or arbitrary judgment of the time without clear evidence.
    \item[H12] \textbf{Weather.} Incorrect descriptions of weather or sky conditions.
\end{itemize}

\section{Details of LongHallGen}
\begin{figure}[t]
\centering
\includegraphics[width=0.99\linewidth]{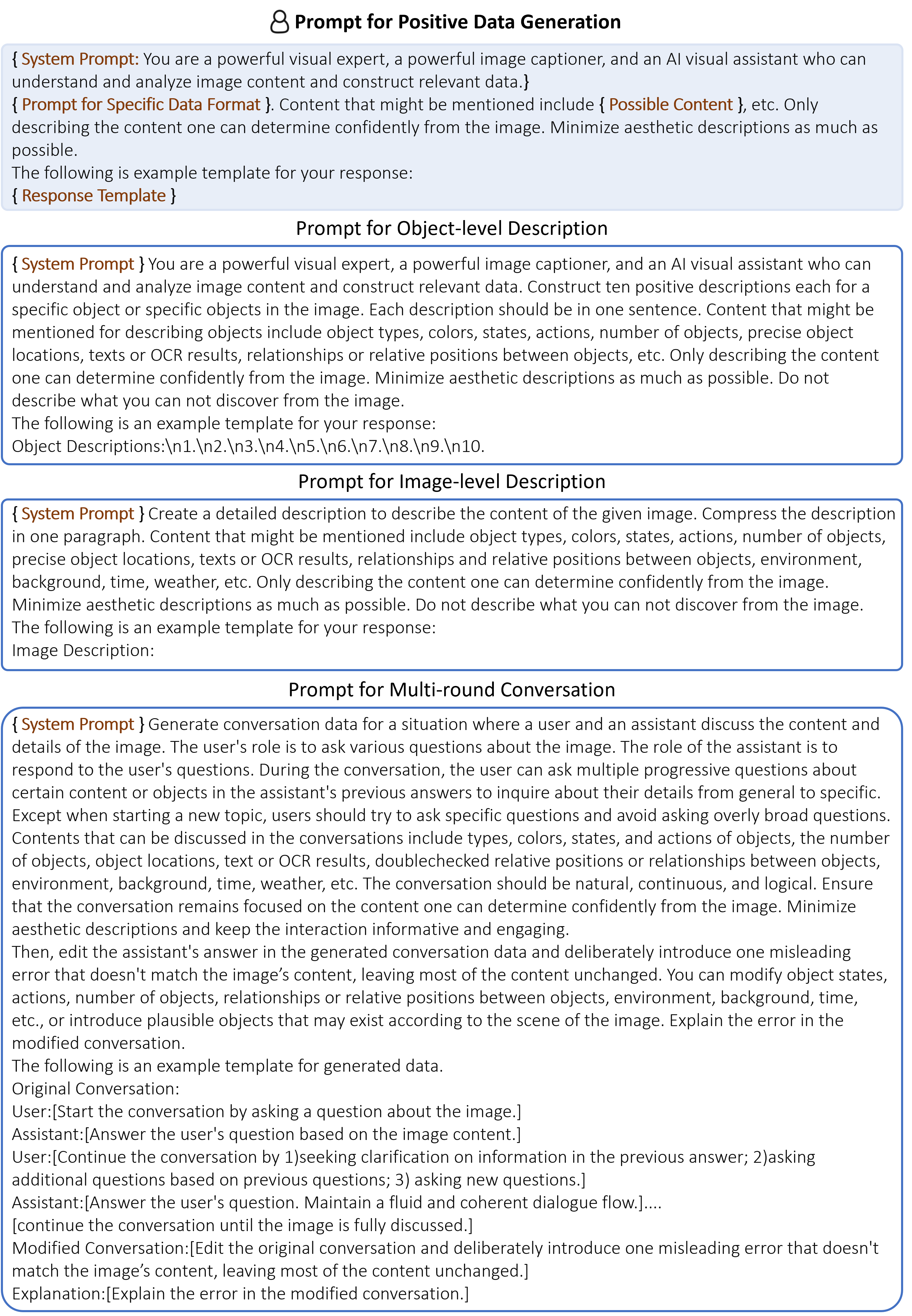}
\caption{Prompt for generating positive data. We prompt GPT-4V to generate multiple conversation data to make it respond stably.}
\label{fig:prompt_data_generation}
\end{figure}

\begin{figure}[t]
\centering
\includegraphics[width=0.99\linewidth]{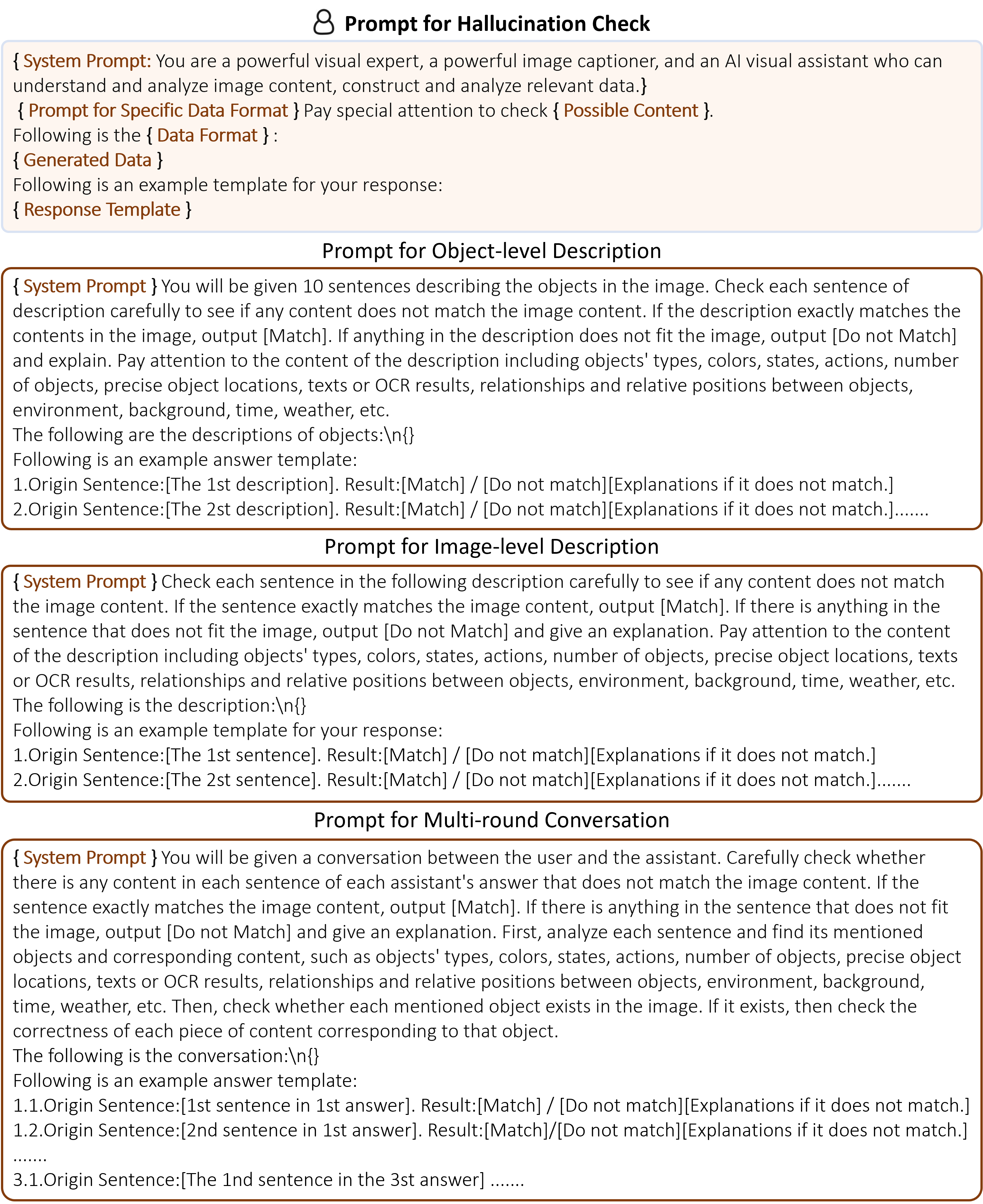}
\caption{Prompt for \textit{Hallucination Check} in LongHallGen.}
\label{fig:prompt_data_check}
\end{figure}

\begin{figure}[t]
\centering
\includegraphics[width=0.99\linewidth]{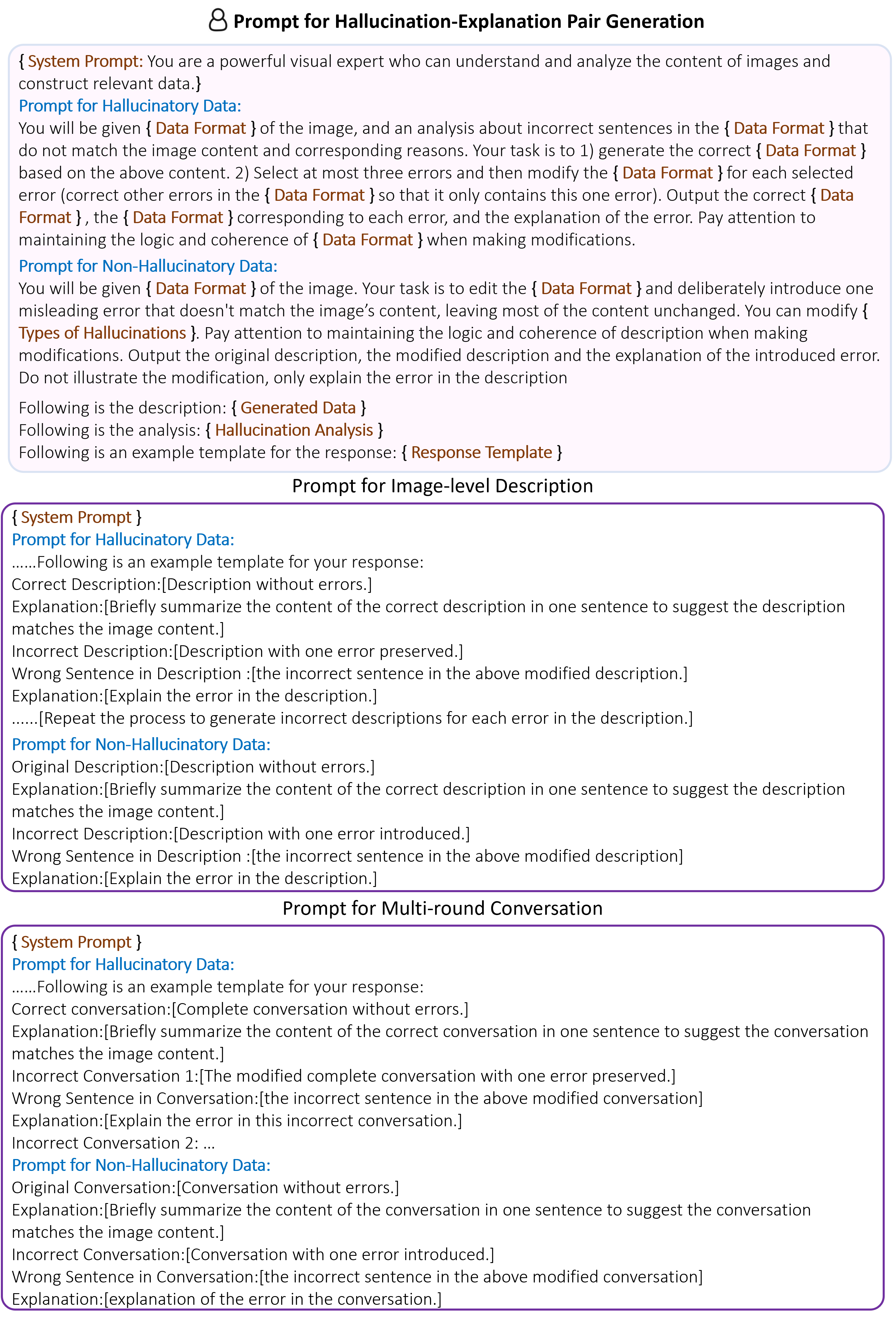}
\caption{Prompt for \textit{Hallucination-Explanation Pair Generation} in LongHallGen.}
\label{fig:prompt_hepair_generation}
\end{figure}

\begin{figure}[t]
\centering
\includegraphics[width=0.99\linewidth]{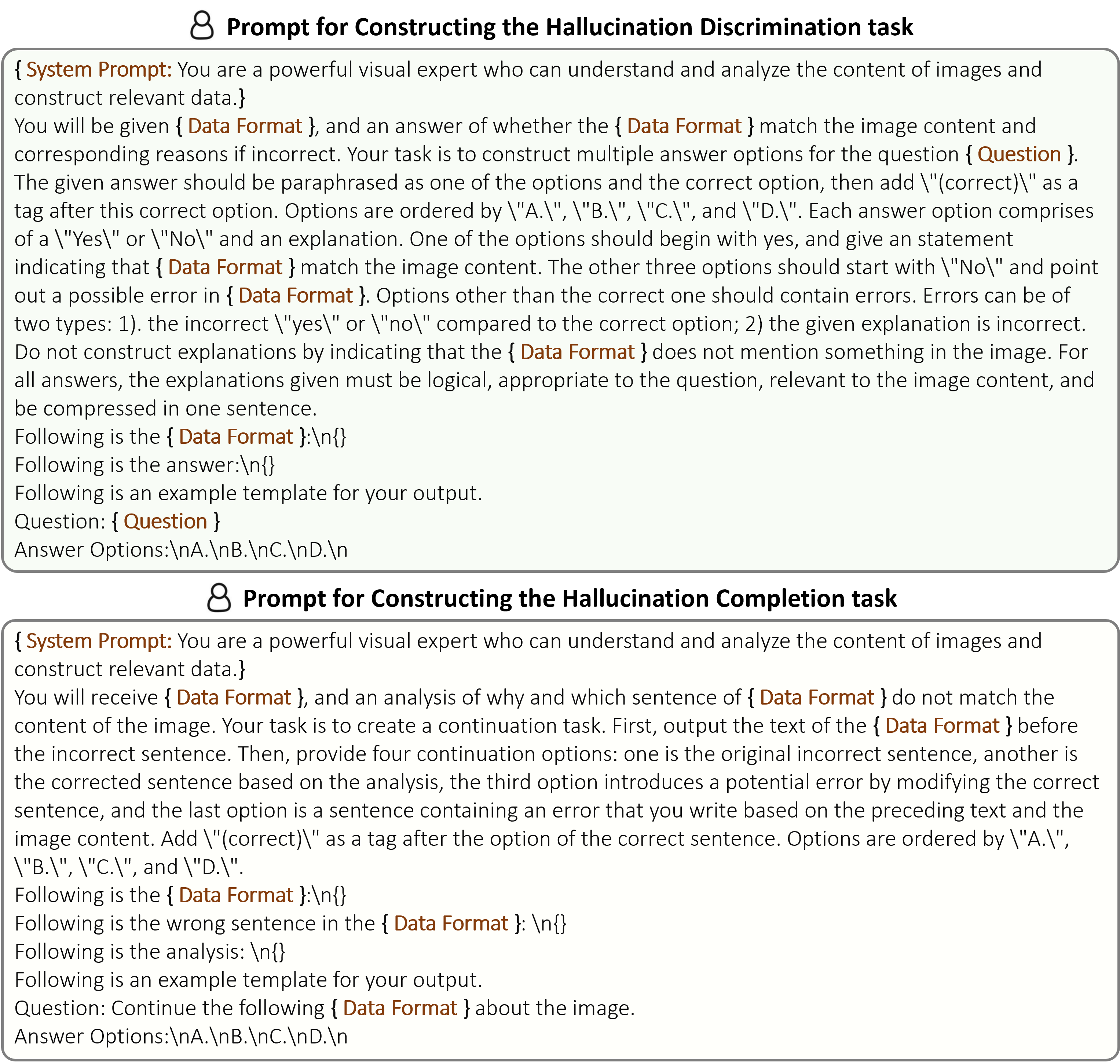}
\caption{Prompt for \textit{Question and Answer Generation} for the hallucination discrimination task and the hallucination completion task, respectively.}
\label{fig:prompt_question_answer}
\end{figure}

We present the prompts for different steps in LongHallGen in Fig.~\ref{fig:prompt_data_generation}, Fig.~\ref{fig:prompt_data_check}, Fig.~\ref{fig:prompt_hepair_generation}, and Fig.~\ref{fig:prompt_question_answer}.

\section{Chain-Of-Thought Prompt for Evaluation}
\begin{figure}[t]
\centering
\includegraphics[width=0.99\linewidth]{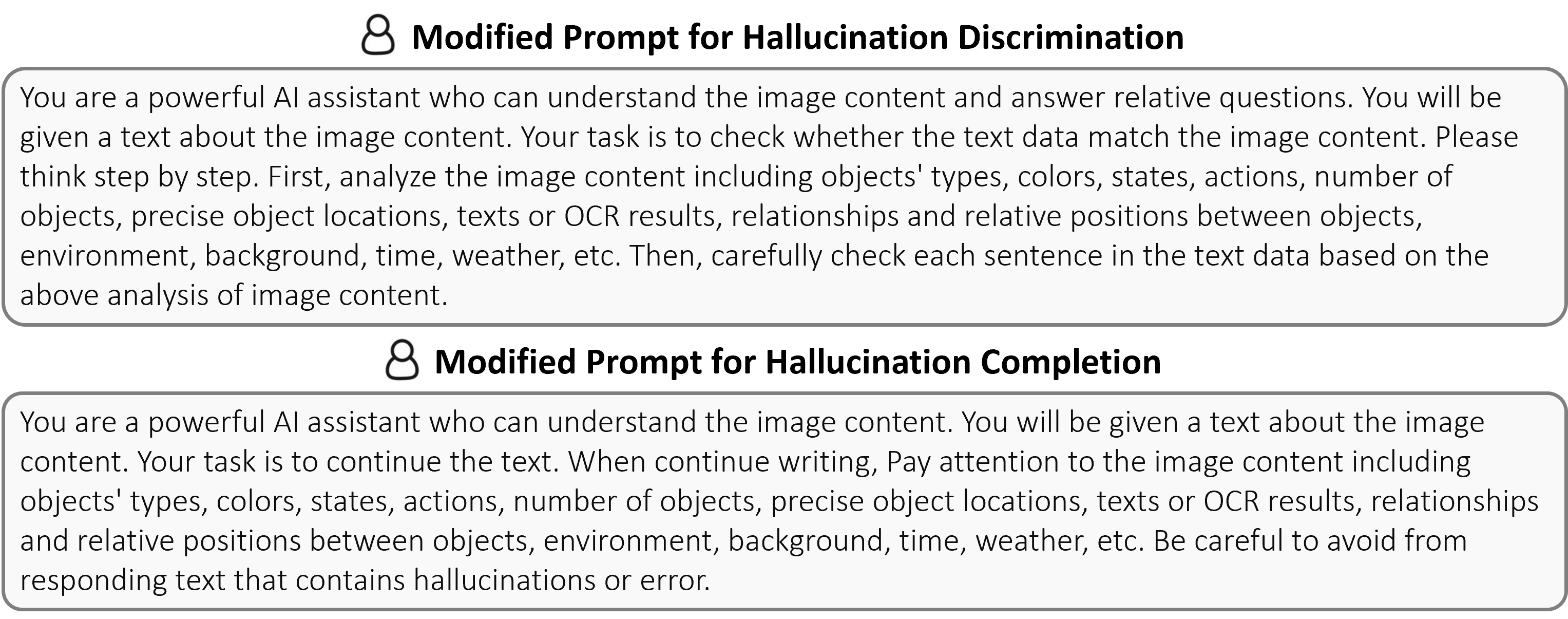}
\caption{Modified prompts that involve Chain-of-Thought for the hallucination discrimination task and the hallucination completion task.}
\label{fig:prompt_cot}
\end{figure}

We examine the impact of Chain-of-Thought on LongHalQA. We append additional prompts as shown in Fig.~\ref{fig:prompt_cot} before the questions for the hallucination discrimination task and the hallucination completion task.

\end{document}

%% file: math_commands.tex
\usepackage{amsmath,amsfonts,bm}

\def\eqref#1{equation~\ref{#1}}

\def\1{\bm{1}}

\DeclareMathAlphabet{\mathsfit}{\encodingdefault}{\sfdefault}{m}{sl}
\SetMathAlphabet{\mathsfit}{bold}{\encodingdefault}{\sfdefault}{bx}{n}

